\documentclass{article}

\usepackage{microtype}
\usepackage{graphicx}
\usepackage{subfigure}
\usepackage{booktabs} 

\usepackage{amsmath}
\usepackage{amssymb}
\usepackage{xcolor}
\usepackage{amsthm}

\newtheorem{example}{Example}
\newtheorem{corollary}{Corollary}
\newtheorem{theorem}{Theorem}

\newtheorem{definition}{Definition}

\usepackage[utf8]{inputenc} 
\usepackage[T1]{fontenc}    
\usepackage{hyperref}       
\usepackage{url}            
\usepackage{booktabs}       
\usepackage{amsfonts}       
\usepackage{nicefrac}       
\usepackage{microtype}      
\usepackage{lipsum}
\usepackage{amssymb}
\usepackage{amsmath}
\usepackage{graphicx}
\usepackage{listings}
\usepackage[toc,page]{appendix}
\usepackage{color}
\usepackage[normalem]{ulem}
\DeclareMathOperator*{\argmax}{arg\,max}

\usepackage{tikz}
\usetikzlibrary{fit,positioning}
\usetikzlibrary{bayesnet}
\usetikzlibrary{arrows}
\usepackage{color}
\usepackage{graphicx}
\usepackage{caption}
\usetikzlibrary{backgrounds}
\usepackage[short]{optidef}
\usepackage{amsthm}
\usepackage{amsmath}
\usetikzlibrary{matrix}
\usepackage{ dsfont }

\usepackage{hyperref}

\usepackage[accepted]{icml2020}

\icmltitlerunning{Multi-Principal Assistance Games}

\begin{document}

\twocolumn[
\icmltitle{Multi-Principal Assistance Games}



\icmlsetsymbol{equal}{*}

\begin{icmlauthorlist}
\icmlauthor{Arnaud Fickinger}{to}
\icmlauthor{Simon Zhuang}{to}
\icmlauthor{Dylan Hadfield-Menell}{to}
\icmlauthor{Stuart Russell}{to}
\end{icmlauthorlist}

\icmlaffiliation{to}{Department of EECS, University of California, Berkeley, USA}

\icmlcorrespondingauthor{Arnaud Fickinger}{arnaud.fickinger@berkeley.edu}

\vskip 0.3in
]



\printAffiliationsAndNotice{}  

\begin{abstract}
Assistance games (also known as cooperative inverse reinforcement learning games) have been proposed as a model for beneficial AI, wherein a robotic agent must act on behalf of a human principal but is initially uncertain about the human’s payoff function. This paper studies multi-principal assistance games, which cover the more general case in which the robot acts on behalf of N humans who may have widely differing payoffs. Impossibility theorems in social choice theory and voting theory can be applied to such games, suggesting that strategic behavior by the human principals may complicate the robot’s task in learning their payoffs. We analyze in particular a “bandit apprentice” game in which the humans act first to demonstrate their individual preferences for the arms and then the robot acts to maximize the sum of human payoffs. We explore the extent to which the cost of choosing suboptimal arms reduces the incentive to mislead---a form of natural mechanism design. In this context we propose a social choice method that uses shared control of a system to combine preference inference with social welfare optimization.
\end{abstract}
\section{Introduction}

The growing presence of AI systems that collaborate and coexist with humans in society highlights the emerging need to ensure that the actions of AI systems benefit society as a whole. This question is formalized as the \textit{value alignment} problem in the AI safety literature \cite{amodei2016concrete}, which emphasizes the need to align the increasingly powerful and autonomous systems with those of their human principal(s). However, humans are prone to misspecify their objectives which can lead to unexpected behaviors \cite{amodei2016concrete}; hence research in value alignment has focused on deriving preferences from human actions. In the body of research in value alignment and human robot interaction, the majority of the work involves scenarios with one human and one AI system. It is an appealing setting because the robot and the human share the same goal. Therefore, methods in this setting such as inverse reinforcement learning \cite{ng2000algorithms, abbeel2004apprenticeship, ramachandran2007bayesian}, inverse reward design \cite{hadfield2017inverse}, and LILA \cite{DBLP:journals/corr/abs-1906-10187} revolve around how an AI system can optimally learn the preferences of the human and apply these results to novel environments. Similarly, the human's incentive is to optimally teach the robot its own preferences. The combination of a learning AI system and a teaching human yields the \emph{assistance game} (also known as the cooperative inverse reinforcement learning game) \cite{hadfield2016cooperative}.

However, AI systems in the real world do not fit this one human, one AI paradigm. Recommendation systems, autonomous vehicles, and parole algorithms do not exist in a vacuum---they often influence and are influenced by multiple humans. Hence, we consider a variation on assistance games where a robot acts on behalf of multiple humans, which we call the \emph{multi-principal assistance game} (MPAG). The key difference between this and the scenario with only one human is that, in general, different humans have different preferences, so it is impossible to align the AI to perfectly match the preferences of everyone. The problem of aggregating individual preferences for making collective decisions has been studied by economists and philosophers for more than two hundred years and constitutes the heart of social choice theory \cite{sen1986social}. 

Even with a given aggregation method, however, the inference process itself is challenged by the presence of selfish agents. While the robot acts to optimize the aggregate of utilities, each human acts to optimize their own utility. Therefore, unlike the single-principal assistance game, the multi-principal assistance game is no longer fully cooperative. This creates a problem for existing value alignment algorithms. These algorithms work under the assumption that the demonstrations and information provided are truly representative of the human's preferences. However, the misalignment between the AI system and each human's preferences yields a perverse incentive for the humans: can they misrepresent their preferences to gain a more desirable outcome? 

We begin with a subclass of MPAGs that generalizes apprenticeship learning. In \textit{multi-principal apprenticeship learning}, the robot observes trajectories from multiple humans and then produces a trajectory that maximizes a social aggregate of the inferred rewards. We state an impossibility result for this setting based on Gibbard's theorem in social choice theory. Our experiments confirm that human demonstrations may indeed ``misrepresent'' their preferences, given a robot that runs maximum entropy inverse reinforcement learning.

We contrast the impossibility result by introducing another subclass of MPAGs based on the multi-armed bandit setting. In the \textit{multi-principal bandit apprentice} setting, the robot is teleoperated by multiple humans. We show under this setting that because demonstrations yield an immediate reward, learning from demonstrations can decrease the incentive to misrepresent one's preferences by incurring a cost of lying. By drawing an analogy between our setting and voting theory, we bring a new perspective on the impossibility results by showing that voting by \textit{demonstrating} reduces the proportion of manipulable profiles. In this setting, the robot can choose which human to give control to and whether to perform an action or not. We use this active learning as a basis to construct an approximately efficient mechanism where humans are incentivized to share the full spectrum of their preferences.



\subsection{Related Work}
\paragraph{Value Alignment.}  
The need for AI systems to align with the preferences of humans  is well documented in AI safety literature \cite{amodei2016concrete}. A first line of work formulates goal inference as an inverse planning problem \cite{baker2007goal}. For example, Inverse Reinforcement Learning computes a reward such that the observed trajectory is optimal in the underlying Markov Decision Process (MDP) \cite{ng2000algorithms} \cite{ziebart2008maximum}. A common assumption of inverse planning methods is that the robot does not influence the decision-making of the human. However, previous work has shown that the presence of a robot has a significant influence on humans \cite{robins2004effects} \cite{kanda2004interactive}. Furthermore, it has been shown that the robot can benefit from interacting with the human to infer the goal. For example, Hadfield-Menell et al. have shown that if we formulate goal inference as a game between the human and the robot, observing the optimal trajectory of the human is in general a sub-optimal strategy \cite{hadfield2016cooperative}. On the contrary, previous work has experimentally shown the emergence of active learning and teaching when optimizing for a joint policy in the value alignment problem \cite{DBLP:journals/corr/abs-1906-10187}. Therefore, modelling collaboration as a game, where both human and robot are aware of their mutual influence, is arguably the most promising approach for efficient human-robot interaction (HRI) \cite{DBLP:journals/corr/Dragan17}.

\paragraph{Mechanism Design.} An important result in social choice theory and mechanism design is the Gibbard–Satterthwaite theorem, which states that, for universal domain of utility functions, every non-trivial game form is subject to strategic or dishonest actions from the players \cite{gibbard1973manipulation}, which can be extended to non-deterministic mechanisms as well \cite{gibbard1978straightforwardness}. This impossibility theorem applies the most general case of multi-principal assistance games as well. Approaches in mechanism design seek to create games where players each acting rationally yield the desired outcome. In a pseudo-linear environment, the VCG mechanism and the expected externality mechanism achieve different forms of incentive-compatibility, meaning players are incentived to act truthfully \cite{borgers2015introduction}. These mechanisms do so by impose transfers, so the externalities of a player's strategic behavior are borne by that player. 

\paragraph{Voting Theory.} Similarly, voting theory, a branch of social choice theory, has also focused on building systems robust to human manipulation. Recently much attention has been given to incorporating ideas of Voting Theory in the design of multiagent systems \cite{ephrati1996deriving}. Our work formalize voting theory in a hybrid human-robot setting: humans ``vote" via their demonstrations, and the robot's resulting actions represent the resulting ``collective decision." In particular, our setup is similar to ordinal voting, since the robot does not access to a cardinal utility function \cite{BOUTILIER2015190}; as a result, the collective decision may not be socially optimal \cite{procaccia2006distortion}.

\paragraph{Human-Robot Team} Robot evolving in a multi-human environment has already been studied by works on human-robot team in the HRI literature. Much work has focused on trust building and resource allocation \cite{claure2019reinforcement}. A common assumption is that the robot and the humans have a common payoff known to the robot. Our work generalize this setting to general-sum payoffs possibly unknown to the robot.



\section{Impossibility Result for Learning from Multiple Humans}

\subsection{Multi-Principal Apprenticeship Learning}


We formalize the problem of learning from multiple humans as \emph{multi-principal apprenticeship learning} (MPAL), a specific multi-principal assistance game that elucidates the process of learning from human demonstrations. A MPAL consists of a multi-agent \textit{world model}, a Markov decision process without a reward function, $M\backslash R=\langle S,A,P,\mu_0,T\rangle$ \cite{abbeel2004apprenticeship} with $N$ humans and one robot where:
\begin{itemize}
    \item $S$ is the set of states.
    \item $A$ is the set of actions.
    \item $P: S \times A \times S \rightarrow [0,1]$ is the transition function.
    \item $\mu_0$ is the initial state distribution
    \item $T$ is the horizon
\end{itemize}


Each human $h$ has a private reward function $R^*_h:S\rightarrow \mathbb{R}$ that is unknown to the robot. We use a \emph{social welfare function} $W$ to aggregate these individual preferences into a single objective $R^* = W(R^*_1,\ldots,R^*_N)$. The robot's objective is to maximize $R^*$ in the world model defined above, despite initial uncertainty about the individual rewards $R^*_i$. Social welfare functions are a heavily studied field, examples include the \textit{utilitarian} criterion $W_U(R^*_1,...,R^*_N) = \sum_h R^*_h$ \cite{6918520} and the \textit{egalitarian} criterion $W_E(R^*_1,...,R^*_N) = \min_h R^*_h$ \cite{zhang2014fairness,nace2008max}.

The robot doesn't have direct access to the human reward functions. Instead each human $h$ provides a collection of $p_h$ trajectories through the state space $\xi^h = (\xi^i_1,...,\xi^i_{p_i}) \in \Xi^{p_h}$, where $\Xi= (S \times A)^{T-1} \times S$. Each trajectory is drawn from $h$'s policy, $\psi_h \in \triangle \Pi$, where $\Pi$ denote the set of deterministic policies in this MDP. Therefore, our overall objective is to build a \textit{mechanism} $\mathcal{M}: (\Xi)^{\sum p_h} \rightarrow \triangle \Pi$ such that $\mathcal{M}(\xi^1,...,\xi^N)$ is optimal in $M^*$.

\begin{example}[MPAL via IRL]
One such mechanism leverages inverse reinforcement learning~\cite{abbeel2004apprenticeship}. This method estimates each reward separately and optimizes the robot's policy for the estimated aggregation of rewards. Formally:
\begin{itemize}
    \item $IRL: \mathcal{P}(\Xi) \rightarrow \mathbb{R}^S$ defined on the set of subsets of $\Xi$
    \item $RL: \mathbb{R}^S \rightarrow \Pi$ returns an optimal policy
    \item $\mathcal{M}(\xi^1,...,\xi^N) = RL \circ W (IRL(\xi^1),...,IRL(\xi^N)) $
\end{itemize}
\end{example}

Note that this formalism also accounts for $IRL$ methods that return a distribution over rewards because we can always marginalize over the uncertainty in the reward function~\cite{ramachandran2007bayesian}: 

\begin{theorem}
If $IRL$ returns a distribution over reward, then the mechanism defined by $\mathcal{M}(\xi^1,...,\xi^N) = RL \circ \mathbb{E}[W(IRL(\xi^1),...,IRL(\xi^N))]$ maximizes the expected value function over the induced distribution of MDPs.
\end{theorem}

\begin{example}[Voting]
If the world model is stateless and the reward functions are defined on the action space, then MPAL is a voting system where humans get immediate reward by voting.
\end{example}



\subsection{Manipulability of Multi-Principal Apprenticeship Learning}
\label{sec:manip-MPAL}
We assume in this section that the mechanism $\mathcal{M}$ is defined directly on the space of strategies of the humans $(\triangle \Pi)^N$.

The humans receive a reward when performing a demonstration (\textit{learning phase} \cite{hadfield2016cooperative}) and a reward when the robot acts in the MDP (\textit{deployement phase}). Therefore, the total expected utility for $h$ is the combination of the two phases:
\begin{equation}
\begin{aligned}
U_h(\psi_h, \psi_{-h}, p_i, \mathcal{M}) = & p_i \mathbb{E}_{\pi_h \sim \psi_h}(V^{\pi_h}(R^*_h)) + \\
& \mathbb{E}_{\pi_r \sim \mathcal{M}(\psi)}(V^{\pi_r}(R^*_h))
\end{aligned}
\end{equation}
where $V^{\pi}(R)$ is the value of the policy $\pi$ in the MDP induced by $R$.

More generally, we introduce a coefficient $\alpha$ that quantify the relative weight that the humans put on the learning phase:

\begin{equation}
\begin{aligned}
U_h(\psi_h, \psi_{-h}, \alpha_i, \mathcal{M}) = & \alpha_i \mathbb{E}_{\pi_h \sim \psi_h}(V^{\pi_i}(R^*_h)) + \\ & (1-\alpha_i)\mathbb{E}_{\pi_r \sim \mathcal{M}(\psi)}(V^{\pi_r}(R^*_h))
\end{aligned}
\end{equation}

In our case, the more demonstrations a human provides, the more weight they will put on the learning phase. In other words, $\alpha$ increases as the number of demonstrations increases. In the contrary, $\alpha$ decreases as the number of times the robot acts increases.

The total utility each human gets depends on the strategy of the other players through the robot's inference. Therefore, even if the humans act independently in the learning phase, the shared interest in the robot's actions during deployment induces a game between them. 

Ideally we would like to have a mechanism such that the action of one human is not influenced by the action of other humans. This would ensure that the mechanism is not manipulable and stays aligned with its initial purpose. Formally:

\begin{definition}[Straightforward Mechanism]
We say that $\mathcal{(M,\alpha)}$ is straightforward if every human $H_i$ has a dominant strategy in the game induced by $\mathcal{M}$ and $\alpha$:
\begin{equation}
\begin{aligned}
\forall h \in [1,N], \exists \psi^*_h \forall \psi_{-h}, \forall \psi_{h}, & U_h(\psi^*_h, \psi_{-h}, \alpha_h, \mathcal{M})\geq \\ & U_h(\psi_h, \psi_{-h}, \alpha_h, \mathcal{M})
\end{aligned}
\end{equation}
\end{definition}

When $\alpha$ is small, i.e. the demonstration is relatively insignificant compared to the robot actions, the Gibbard–Satterthwaite \cite{gibbard1973manipulation} can be applied to show that the only straightforward mechanism are trivial.
\begin{theorem} [Based on Gibbard 1973]
For sufficiently small $\alpha$, the only straightforward deterministic mechanisms are as follows:
\begin{itemize}
\item Duple mechanisms, where the set of possible trajectories are restricted to two.
\item There exists one human that can choose among the possible trajectories (dictatorship).
\end{itemize}
\end{theorem}
Furthermore, we can extend Gibbard's 1978 theorem \cite{gibbard1978straightforwardness} for non-deterministic mechanisms.
\begin{theorem} [Based on Gibbard 1978]
	On the domain of versatile\footnote{A strategy is versatile if the set of utility profile for which it is dominant has interior points.} policies, any straightforward mechanism must be a probability mixture of mechanisms of two kind:
	\begin{itemize}
		\item Duple mechanisms
		\item Unilateral games, where one human gets to choose among a certain set of possible lotteries over trajectories.
	\end{itemize}
	\end{theorem}

\subsection{Experiment: Attacking Inverse Reinforcement Learning}
The theorems in Section~\ref{sec:manip-MPAL} apply when $\alpha$ is sufficiently small. However, there can still be incentives for strategic behavior in games where $\alpha$ is non-negligible. In this section, we consider a mechanism based on Maximum Entropy IRL~\cite{ziebart2008maximum} and introduce a solver to manipulate it.

More specifically, our mechanism has 3 steps. First, we aggregates all of the human player's trajectories into a single dataset. Second, the mechanism uses Maximum Entropy IRL~\cite{ziebart2008maximum} to infer a reward reward function. Finally, we execute a policy that optimizes this reward function. Formally: 
\begin{equation}
\begin{aligned}
\mathcal{M}(\xi^1,...,\xi^N) = RL \circ MEIRL(\xi^1,...,\xi^N)
\end{aligned}
\end{equation}

\begin{theorem}
The mechanism presented above is not straightforward.
\end{theorem}

To show that, we introduce a quadratic program (QP) solver that heuristically creates adversarial trajectories against Maximum Entropy IRL. Similarly to previous work on single-agent value alignment \cite{hadfield2016cooperative, NIPS2016_6413}, the QP solver finds an approximate best-response trajectory in a three-player game with one robot and two humans. We suppose that each human gives a single trajectory to the robot and the robot aggregates the trajectories to find a single reward parameter to train its policy.

The human's goal is to optimize for immediate reward, balanced with future reward from the robot's deployment. While it is hard to directly optimize the result of the robot's inference, the average feature counts in the trajectory dataset have been used as an effective proxy\cite{hadfield2016cooperative, NIPS2016_6413}. Formally, we capture this by defining a QP that optimizes for a combination of immediate reward and the distance of the final features from a target (see appendix for the full derivation):
\begin{equation}
\begin{aligned}
\max_{\rho^t_{s, a}} & \sum_{s, a, t} \gamma^t \rho^t_{s, a} \phi(s)^T w \\
&- \lambda||\sum_{s, a, t} \rho^t_{s, a} \phi(s)-  (2\mathbb{E}[\phi | w]-\phi(\xi^1))||^2 \\
\text{s.t} & \sum_{a} \rho^{t+1}_{s, a} =  \sum_{s', a} P(s', a, s) \rho^t_{s', a} \\
& \forall s, \forall t\in [0, T-1]\\
&\sum_{a} \rho^0_{s, a} = \mu_0[s] \, \, \forall s
\end{aligned}
\end{equation}

where $\rho$ is the occupancy measure, $\phi$ is the feature space embedding, $P$ is the transition matrix, $T$ is the horizon, $\mu_0$ is the initial state distribution, $\lambda$ weights the relative importance of the outcome of the robot's policy, $\omega$ is the reward parameter, $\mathbb{E}[\phi | w]$ is the expected feature count of a policy optimal for the MDP induced by $\omega$ and $\xi^1$ is the trajectory of the first human.

This is a regularized dual of the linear program formulation for finite-horizon discounted Markov Decision Process \cite{puterman2014markov}. The best-response trajectory can be directly derived from the occupancy measure.

The experimental results in a 2D gridworld environment are presented in Figure 1. The environment is characterized by a three-dimensional feature space and an horizon of 40. The reward parameter of the second human is fixed and equal to $\omega_2 = (0.9, 1, 0)$. We compute their best-response trajectories to two different humans, one with reward parameter $\omega_1 = (1, 0, 0)$ and another with reward parameter $\omega_1 = (0, 0, 1)$. In the former case, the total utility is maximized by playing the optimal trajectory but in the latter case, the best-response found by our QP solver is not the optimal trajectory. Figure 1 presents the optimal trajectory versus approximate best response in the latter case.


\begin{figure}[h]
\centering
\includegraphics[width=0.5\textwidth]{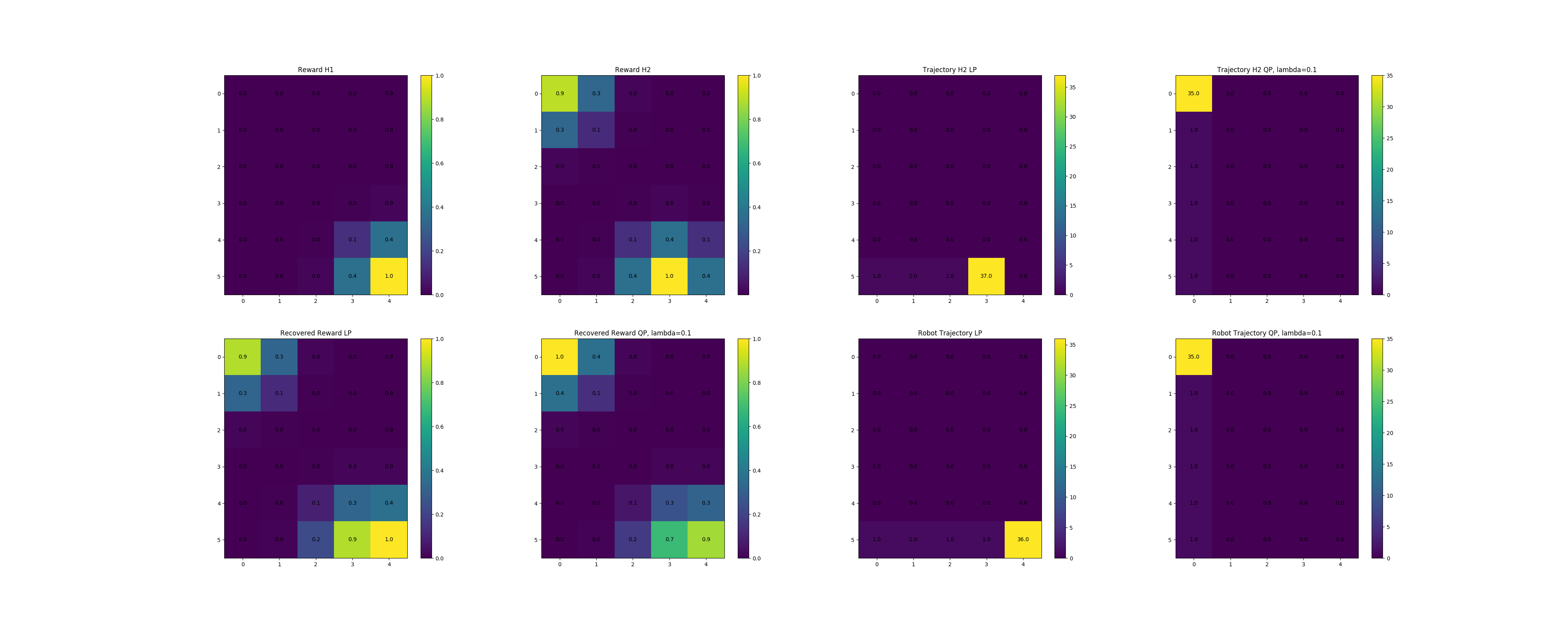}
\caption{Manipulating a Multi-Agent Alignment IRL Method using a QP in a 2D $5 \times 6$ Gridworld Environment with a 3D feature space.  First row: True reward of humans 1 and 2; State visitation count of optimal (resp. best-response) trajectories of human 2 (the initial state is in the bottom left-hand corner). Second row: Recovered rewards using IRL on the aggregate of first human's optimal and second human's optimal (resp. best-response) trajectories; Optimal robot trajectories in the MDP induced by these rewards.}
\end{figure}

\section{Mechanism Design for Multi-Agent Alignment}
We now propose a social choice method that uses shared control of a system to combine preference inference with social welfare optimization. In this context, we demonstrate the possibility of non-trivial straightforward and approximately efficient mechanisms.
\subsection{Multi-Principal Bandit Apprentice (MPBA)}
Imagine a teloperation setting with multiple human principals. The robot wants to implement a policy that will have an impact on several humans. To choose a policy optimized for this specific group of human it needs to learn about each individual's preferences by letting them operate simultaneously or successively. If it let them operate only a few times, it might not get enough information about their preferences and end up with a suboptimal policy in term of social welfare. In the contrary, as long as it is operated by humans, the robot's policy is not optimized for the social welfare but rather each individual's preferences. We model this exploration-exploitation problem with a multi-armed bandit setting adapted to our multi-agent alignment problem.

In the classical setting of the multi-armed bandit, a single player receives an arm-dependent stochastic reward each time they pull an arm. Their goal is to find a policy mapping history of actions and rewards to actions that minimizes the regret by finding the good trade-off between exploration and exploitation. 

We deviate from the classical setting in several respects. First, $N>1$ humans are pulling arms and the rewards on each arm are deterministic, specific to each human, and known to each human $R^*_h:[1,M]\rightarrow[0,1]$. The humans are not exploring; they are communicating information to the robot.
Second, the robot can decide to pull an arm itself or it can choose one human to pull an arm in the next round. Third, when a human pulls an arm, the robot observes only that fact and does not observe the reward received. We assume that each person's total utility is the same: $\forall h, \sum_a R^*_h(a)=1$.

As in the previous part, we suppose that the designers of the system would like to maximize a \emph{social welfare} function that aggregates utility. Formally, we define the social welfare of arm $a$ as $w_a=\frac{1}{N}\sum_{h=1}^NR^*_h(a)$ and $w^* = \max_a w_a$. For $t \in [1,T]$, the random variables $H_t$, $A_t$, and $W_t$ represent respectively the human chosen, the action chosen and the social welfare obtained at time $t$. Since the robot can choose itself, the codomain of $H_t$ is [1,N+1]. We also denote by $\tilde{H}_t$ the restriction of $H_t$ to [1,N]: $p(\tilde{H_t}) = p(H_t|H_t\leq N)$ The objective of the robot is to minimize cumulative regret with respect to social welfare:
\begin{equation}
\begin{aligned}
\min R_T & =  \mathbb{E}\sum_{t=1}^T W^* - W_t
\end{aligned}
\end{equation}

\subsection{Hardness of MPBA}

To begin with we consider the non-strategic setting. We suppose that the humans implement a stationary policy to approximately optimize immediate reward:
\begin{equation}
\begin{aligned}
p(A_t|\tilde{H_t}) \propto e^{\beta R^*_{\tilde{H_t}}(A_t)}
\end{aligned}
\end{equation}
where $\beta$ is a parameter describing how close are humans from making optimal decisions.

A setting similar to ours has been studied to formalize assistance to a single human \cite{DBLP:journals/corr/abs-1901-08654}. In their setting, at each round the single human suggests an arm to pull and the robot pulls an arm based on the arms suggested so far. They show that if the human is noisily optimal---the probability of suggesting the best arm is strictly greater than the probability of suggesting a suboptimal arm---then simply pulling the most commonly suggested arm achieves finite expected regret, contrasting with the lower bound in $\Omega(\log T)$ for the classical setting. 

The following theorems show that the multi-agent setting is harder than its single-agent counterpart.

\begin{theorem}
In a bandit assistance game with a single human, any utility profile leads to zero regret when $\beta \rightarrow \infty$ and the robot uses an explore-then-commit strategy.
\end{theorem}

\begin{theorem}
In a bandit assistance game with multiple humans, there is a utility profile such that the regret is in $\Omega(T)$ when $\beta \rightarrow \infty$ and the robot uses an explore-then-commit strategy.
\end{theorem}

Intuitively, inferring the best arm is not sufficient to maximize the social welfare when there are multiple humans.

\subsection{Incentive-Compatibility of Voting by Demonstrating}
We now consider the strategic setting. Each human $h$ acts following a policy $\psi_h:([1,M] \times [1,N])^*\rightarrow\triangle[1,M]$ mapping history of human-action pairs to action and aims to maximize its utility $\mathbb{E}(\sum_t r_h^*(A_t))$.

MPBA can be seen as a voting system where instead of \textit{announcing} their type, the humans \textit{demonstrate} it. The next theorems show that our setting is more robust to manipulation than classical voting systems. 

Formally, define the truthfulness $\gamma_h$ of a human strategy $\psi_h$ as the frequency of choosing among the best arms:
\begin{equation}
\begin{aligned}
\gamma_h = \frac {\sum_1^T \mathds{I}(H_t=h)\mathds{I}(A_t \in \argmax_a R^*_h(a))}{\sum_1^T \mathds{I}(H_t=h)}
\end{aligned}
\end{equation}
We say that $\psi_h$ is truthful if $\gamma_h = 1$.

We assume that the robot uses an explore-then-commit strategy with an exploration time of $N \cdot T$ and an exploitation time of 1. The next theorem states that increasing the exploration time decreases the number of non-dominated untruthful strategies. We denote by $\triangle^*_h$ the minimal suboptimal gap of $h$ and we assume that every humans has at least one suboptimal arm.

\begin{theorem} 
Given $\gamma \in ]0,1[$, if $T > \frac{R^*_h}{(1-\gamma)\triangle^*_h}$ then any strategy $\psi_h$ such that $\gamma_h<\gamma$ is strictly dominated by a truthful strategy.
\end{theorem}

\begin{example}[Plurality Voting with Shared Control]
The robot chooses the arm to pull following: $a_R = \argmax_a \sum_{i=1}^N \mathds{1}(\tilde{a}_i=a)$ where $\tilde{a}_i = \argmax_a \sum_{t=1}^T \mathds{1}(a^i_t=a)$.
\end{example}

By using Theorem 4, we can characterize the incentive-compatibility of this mechanism.

\begin{corollary}
For any domain of utilities $\mathcal{D}_{\epsilon,C} = \{u \in \mathbb{R}^M: u^*<C \wedge (u^*=u^{**} \vee u^*-u^{**}>\epsilon)\}$, where $u^* = \max_a u(a)$ and $u^{**}=\max_a\{u(a): u(a)\neq u^*\}$, if $T>\frac{2C}{\epsilon}$, then the plurality voting with shared control is non-dictatorial, does not limit the possible outcomes to two alternatives and it is dominant-strategy incentive-compatible on $\mathcal{D_{\epsilon, C}}$. In the limit $T \rightarrow \infty$ we have an incentive-compatible mechanism on the universal domain.
\end{corollary}


When the exploration time is equal to the exploitation time, voting by demonstrating is subject to manipulation. We can nevertheless quantify the robustness of such a system by comparing its proportion of \textit{manipulable profiles} to the one of classical systems. Formally, let's define a manipulable profile:

\begin{definition}[Manipulable profile]
We say that a profile $r_h^*$ is manipulable in the game induced by $\mathcal{M}$ if there is $\psi_{-h}$ and $\psi_h$ such that $\gamma_h<1$ and for any truthful strategy $\psi^*_h$, $U_h(\psi^*_h, \psi_{-h}, \mathcal{M})<U_h(\psi_h, \psi_{-h}, \mathcal{M})$.
\end{definition}

Using a geometric argument on the 2-simplex we can prove the following:

\begin{theorem}
In a system using plurality voting with random tiebreak with 3 voters and 3 alternatives, the set of manipulable profile by demonstrating is included in the set of manipulable profile by announcing. Furthermore, the proportion of manipulable profile by announcing but not by demonstrating is $\frac{1}{9}$. 
\end{theorem}

\subsection{Efficient MPBA}
In the previous section we have seen that voting by demonstrating provides a naturally incentive-compatible mechanism. Yet our first aim is to maximize the social welfare, therefore we want to build an approximately efficient mechanism. As we have seen, it is hard to optimize social welfare because observations of optimal human behavior provides limited information the corresponding utility functions. An alternative is to build a mechanism that incentivizes the humans to provide information about their entire utility function, not just their optimal arm. 
By analogy with the voting theory literature, we define the distortion for a given robot strategy $\mathcal{M}$ as the ratio between the optimal social welfare and the social welfare obtained by the robot in the worst case.
\begin{equation}
\begin{aligned}
\triangle(\mathcal{M}) = \max_{R^*}\frac{\max_a \sum_h R^*_h(a)}{\mathbb{E}(\sum_h(R^*_h(a_R^{\mathcal{M}}(\psi^*(R^*, \mathcal{M}))))}
\end{aligned}
\end{equation}
where $\psi^*(R^*, \mathcal{M})$ is the best response to $\mathcal{M}$ for the profile $R^*$.

By building on recent results in ordinal voting theory \cite{BOUTILIER2015190} we can construct an approximately efficient mechanism $\mathcal{M}$. In broad outline, the robot chooses a human and execute the action only if the human did not choose this action before. Periodically, the robot choose a random action with probability $\frac{1}{2^{\frac{1}{M}}}$. We present the full algorithm in the appendix and show that this algorithm incentivizes the humans to share their true ordinal preferences. We derive the following upper-bound for the distortion:

\begin{theorem} 
$\triangle(\mathcal{M}) = O(\sqrt{M\log M})$
\end{theorem}

\section{Conclusion and Future Work}
In this paper, we explore an area of concern in the study of AI alignment---ensuring that AI systems are designed so that humans agents are incentivized to interact with AI systems in a ``honest" way. Applying the Gibbard–Satterthwaite theorem to this scenario indicate that demonstrations with little to no significance are subject to strategic behavior. Experimental results show that a commonly used inverse reinforcement learning paradigm, which works well in single-human alignment instances, is prone to manipulative behavior. However, on a modified setting, we find effective mechanisms can arise from learning human preferences via their actions if those actions are sufficiently consequential.

The overall problem of preventing manipulative behavior in multi-human AI systems is open and presents many opportunities for further work. Our methods are applied to fairly simple problems: there exists a need to generalize these results to more general theoretical settings and more complicated situations in the real world.


\bibliography{final}
\bibliographystyle{icml2020}

\end{document}